\documentclass[journal]{IEEEtai}
\usepackage{url}
\usepackage[utf8]{inputenc}
\usepackage[small]{caption}
\usepackage{graphicx}
\usepackage{booktabs}
\usepackage{algorithm}
\usepackage{cite}
\usepackage{algorithmic}
\usepackage{multirow}
\usepackage{bm}

\usepackage{booktabs}
\usepackage{algorithm,algorithmic}

\usepackage{subfigure}
\usepackage{amsmath,amssymb, amsthm,amsfonts}

\newtheorem{lemma}{Lemma}

\ifCLASSINFOpdf
\else
\fi
\begin{document}
%
\title{Double Self-weighted Multi-view Clustering via Adaptive View Fusion}
%
%
%

\author{Xiang~Fang*,
        Yuchong~Hu,~\IEEEmembership{Member,~IEEE}
\thanks{Corresponding author: Xiang Fang}
\thanks{X. Fang is with the School of Computer Science and Technology, Key
Laboratory of Information Storage System Ministry of Education of China,
Huazhong University of Science and Technology, Wuhan 430074, China (e-mail: xfang9508@gmail.com).}
\thanks{Y. Hu is with the School of Computer Science and Technology, Key
Laboratory of Information Storage System Ministry of Education of China,
Huazhong University of Science and Technology, Wuhan 430074, China (e-mail: yuchonghu@hust.edu.cn).}}

\maketitle


\begin{abstract}
Multi-view clustering has been applied in many real-world applications where original data often contain noises. Some graph-based multi-view clustering methods have been proposed to try to reduce the negative influence of noises. However, previous graph-based multi-view clustering methods treat all features equally even if there are redundant features or noises, which is obviously unreasonable. In this paper, we propose a novel multi-view clustering framework Double Self-weighted Multi-view Clustering (DSMC) to overcome the aforementioned deficiency. DSMC performs \emph{double} self-weighted operations to remove redundant features and noises from each graph, thereby obtaining robust graphs. For the first self-weighted operation, it assigns different weights to different features by introducing an adaptive weight matrix, which can reinforce the role of the important features in the joint representation and make each graph robust. For the second self-weighting operation, it weights different graphs by imposing an adaptive weight factor, which can assign larger weights to more robust graphs. Furthermore, by designing an \emph{adaptive} multiple graphs fusion, we can fuse the features in the different graphs to integrate these graphs for clustering. Experiments on six real-world datasets demonstrate its advantages over other state-of-the-art multi-view clustering methods.
\end{abstract}

%
%
%
%
%
%
%
%
%
%



\ifCLASSOPTIONpeerreview
 \fi
%
\IEEEpeerreviewmaketitle

\section{Introduction}
Spectral clustering can be regarded as the graph-based clustering since its performance is directly determined by the obtained graph. Spectral clustering makes use of the spectral-graph structure of an affinity matrix to partition data into disjoint meaningful groups. In many real-world applications, we often collect multi-view data~\cite{blum1998combining,lu2013unified,Tang2019AAAI}. Therefore, multi-view clustering methods are needed to cluster these multi-view data~\cite{lan2015reducing,gong2019multi,li2018multi}. Benefiting from its efficiency and good performance, multi-view spectral clustering has gained much attention in many fields of machine learning and data mining~\cite{min2017delicious,wang2013collaborative,wang2019compact}. Up to present, several multi-view spectral clustering methods have been proposed~\cite{nie2017multi1,kang2018self}, but most of them perform clustering only based on the diversity between different views in multi-view data~\cite{du2018multi,liu2023exploring,wang2025taylor,fang2026towardsicml,kuai2026dynamic,wang2025point,fang2025your,zhang2025monoattack,fang2023hierarchical,liu2024towards,yang2025eood,fang2022multi,fang2026cogniVerse,lei2025exploring,fang2023you,wang2025dypolyseg,fang2025hierarchical,yan2026fit,fang2025adaptive,wang2026topadapter,cai2025imperceptible,fang2026slap,wang2026reasoning,fang2026immuno,wang2026biologically,fang2026disentangling,wang2025reducing,fang2026advancing,fang2026unveiling,wang2026from,liu2023conditional,liu2026attacking,fang2026rethinking,wang2025seeing,fang2026towards,fang2025multi,fang2024fewer,liu2024pandora,fang2024multi,fang2025turing,fang2024not,liu2023hypotheses,fang2024rethinking,liu2024unsupervised,fang2023annotations,xiong2024rethinking,fang2021unbalanced,wang2025prototype,zhang2025manipulating,fang2026align,tang2024reparameterization,fang2025adaptivetai,tang2025simplification,fang2021animc,cai2026towards,fang2020v}.
%

To learn optimal graphs,  several state-of-the-art multi-view spectral clustering methods are proposed. \cite{nie2017multi1} propose MLAN to learn the local structure of multi-view data by constructing a constrained graph Laplacian. \cite{cai2011heterogeneous} propose MMSC to learn a shared graph Laplacian matrix by integrating different image features.  \cite{xia2014robust} propose RMSC to learn the standard Markov chain by constructing a graph with fewer noises for each view and a low-rank transition probability matrix shared by all the views. \cite{nie2017self} propose SwMC to learn an optimal weight for each graph by constructing a Laplacian rank constrained graph. \cite{nie2018multiview} propose AWP to learn optimal Procrustes~\cite{friedman2001elements} for all the views adaptively.

However, these multi-view spectral clustering methods mainly have the following two pivotal drawbacks which greatly limit their applications: i) they can not learn the intrinsic structure of data because they neglect of the local structure of data; ii) these learned graphs are not the optimal graphs for clustering because they ignore the connection between features in different views. These two drawbacks will cause these methods to fail in many multi-view clustering tasks with a large number of features because these methods are unable to extract favorable features for clustering from high-dimensional features with noises. For convenience, we call these features favorable for clustering "\emph{favorable features}" and these features unfavorable for clustering "\emph{unfavorable features}". 
For the sake of description, we do not distinguish between views and graphs in this paper.

Therefore, to tackle these issues, we try to learn optimal graphs for multi-view clustering by introducing weight matrix to each graph in order to extract favorable features, and these graphs have the following characteristics: 1) each graph corresponds the optimal representation of a view (in the view, these features favorable for clustering are considered and these features unfavorable for clustering are ignored), 2) for all the graphs, these views favorable for clustering are preserved and these views unfavorable for clustering are ignored.

In this paper, we propose a Double Self-weighted Multi-view Clustering (DSMC) scheme to obtain these optimal graphs. DSMC performs two self-weighted operations: a) for the first self-weighted operations, it first creates multiple initialized graphs, then weights the different features of these graphs by introducing weight matrices; b) for the second self-weighted operations, it weights different graphs by learning a weight coefficient. Furthermore, we design an \emph{adaptive} multiple graphs fusion method to integrate these graphs by fusing favorable features, which can reduce the influence of noises.

In summary, we highlight the main contributions of our proposed DSMC as follows:
\begin{itemize}
  \item DSMC is an innovative method to construct optimal graphs. It can weigh these graphs twice and fuses these graphs adaptively to learn optimal graphs for clustering, thus improving clustering results.
  \item Double self-weighted operations are performed to integrate optimal graphs. For the first self-weighted operation, DSMC can remove unfavorable features and extract favorable features by introducing a weight matrix. For the second self-weighted operation, DSMC can integrate these features by learning a suitable weight for each graph.
  \item It fuses these graphs adaptively to simplify the computation of our self-weighted framework and improve clustering performance by redefining the weight of each graph. Experimental results show that it achieves better performance than state-of-the-art multi-view spectral clustering approaches. Empirical comparisons also show the promising efficiencies of DSMC.
\end{itemize}
The rest of the paper is organized as follows. Section~\ref{section:meth} overviews related work, proposes our DSMC method and analyzes it. Section~\ref{section:optimization} leverages an iteration procedure to solve our DSMC. Section~\ref{section:exp} shows the experimental results and analysis. Section~\ref{section:con} concludes the paper.

\textbf{Notation:} For convenience, we introduce some notation through the paper. All the matrices are written as uppercase. For a matrix $\bm{A}$, its $ij$-th element and $i$-th column are denoted by $a_{ij}$ and $a_i$ separately; the trace of $\bm{A}$ is denoted by $\text{Tr}(\bm{A})$; the Frobenius norm of $\bm{A}$ is denoted by $||\bm{A}||_F$; $\bm{E}$ is a matrix which all elements are 1. For the data matrix of one view, it is denoted by $\bm{X}\in \mathbb{R}^{p\times d}$, where $p$ is the number of instances and $d$ is the dimension of features. The weighted graph in spectral clustering is denoted by $\bm{W}$ and $w_{ij}=exp(-\frac{||x_i-x_j||_2^2}{2\sigma^2})$, where $\sigma$ is the bandwidth parameter. $\bm{D}\in \mathbb{R}^{p\times p}$ denotes the diagonal matrix and $d_{ii}=\sum_{j=1}^pw_{ij}$. $\bm{L}$ denotes the normalized graph Laplacian matrix, and $\bm{L}=\bm{I}-\bm{D}^{-1/2}\bm{W}\bm{D}^{-1/2}$, where $\bm{I}$ is an identity matrix. $\bm{F}$ denotes the clustering indicator matrix. $c$ denotes the number of clusters. For a matrix $\bm{A}$, $\bm{A}^{1/2}$ is the element-wise square root of $\bm{A}$, i.e., each element of $\bm{A}^{1/2}$ is $(a_{ij})^{1/2}$. $\odot$ denotes the element-wise multiplication. $k$ denotes the number of clusters. $\bm{1}$ is a column vector and each element of the vector is 1. For a matrix $\bm{X}$, $\bm{X}\in Def$ denotes $\bm{X}\in \{\bm{X}\in\{0,1\}^{p\times k}|\bm{X}\bm{1}=\bm{1}\}$.

%
%

\section{Methodology}   \label{section:meth}
%
In this section, we first revisit some classic methods of multi-view spectral clustering.
\subsection{Multi-view Spectral Clustering Revisit}
\label{subsection:self}
Multi-view spectral clustering is popular in many multi-view clustering tasks for its simplicity and effectiveness. Given an input data matrix $X^{v}\in \mathbb{R}^{p\times d_v}$, and each column of $X^{v}$ is an instance vector, where $d_v$ is the dimension of features in the $v$-th view. To group these instances into $c$ clusters, the classic framework based on multi-view spectral clustering is as follows:
\begin{align}\label{tr}
&\min \sum_{v=1}^nTr(FL^{(v)}F) \nonumber\\
&\mbox{s.t. } F^TF=I
\end{align}
For the framework, a direct way to use each graph obtained from Eq.~\eqref{tr} is to stack up to a new normalized Laplacian matrix and put it into the standard spectral analysis model. But the simple way neglects the differences of different graphs and may add an unreliable graph to Eq.~\eqref{tr}, which will load to unsatisfactory clustering results. To improve the clustering results, \cite{nie2018multiview} propose AWP to learn optimal Procrustes for all the views adaptively, and the framework of AWP is as follows:
\begin{align}\label{lap}
&\min \sum_{v=1}^nw_v||Y-F^{(v)}R^{(v)}||_F^2 \nonumber\\
&\mbox{s.t. } Y \in Def,(R^{(v)})^TR^{(v)}=I.
\end{align}
\subsection{Dual Self-weighted Multi-view Clustering} \label{section:adaptive}
Eq.~\eqref{lap} treats all features (in each view) equally important for clustering. This treatment is harmful to cluster large-scale multi-view data, which always has a large number of unimportant features that are useless for clustering (called \emph{terrible features}), especially in image clustering tasks and text clustering tasks.
However, a robust self-weighted multi-view clustering methods should not only assign optimal weight to each view, but also extract important features to improve clustering results.
Motivated by this, we propose Dual Self-weighted Multi-view Clustering (DSMC) as follows:
\begin{align}\label{jiaw}
&\min \sum_{v=1}^nw_v||(M^{(v)})^{1/2}\odot(Y-F^{(v)}R^{(v)})||_F^2+\frac{\mu}{2}||M^{(v)}||_F^2 \nonumber\\
&\mbox{s.t. } Y \in Def,(R^{(v)})^TR^{(v)}=I,M^{(v)}>0,(M^{(v)})^TE=E,
\end{align}
where $M^{(v)}$ is the weighted matrix, in the $v$-th view, to assign different weights to different features. $(M^{(v)})^{1/2}$ is the element-wise square root of $M^{(v)}$. $\odot$ denotes the element-wise multiplication. $\mu$ is a hyper-parameter to controls the trade-off of corresponding terms, and $E$ is a matrix which all elements are 1.

By introducing weighted matrix $M^{(v)}$ to Eq.~\eqref{lap}, we can reinforce the effect of important features by assigning large weights to them, thus learning robust graphs. Besides, we add the constraint term $(M^{(v)})^TE=E$ to Eq.~\eqref{lap} to treat each instance equally.

Similar to \cite{nie2018multiview}, we can obtain $w_{v}$ as follows:
\begin{align}\label{wgengxin}
w_{v}=\frac{1}{2(||(M^{(v)})^{1/2}\odot(Y-F^{(v)}R^{(v)})||_F)}.
\end{align}
From Eq.~\eqref{jiaw} and Eq.~\eqref{wgengxin}, our DSMC can be formulated as:
\begin{align}\label{xinbanben}
\min\sum\limits_v&w_{v}||(M^{(v)})^{1/2}\odot(Y-F^{(v)}R^{(v)})||_F^2+\frac{\mu}{2}||M^{(v)}||_F^2 \nonumber\\
&w_{v}=\frac{1}{2(||(M^{(v)})^{1/2}\odot(Y-F^{(v)}R^{(v)})||_F)}\nonumber\\
&\mbox{s.t. } Y \in Def,(R^{(v)})^TR^{(v)}=I,S>0,(M^{(v)})^TE=E
\end{align}
Note that $w_{v}$ in Eq.~\eqref{xinbanben} is dependent on the target variable $Y$, so it is not directly available. But we can first set $w_{v}$ stationary, and update it after obtaining $Y$~\cite{nie2018multiview}.

%
%
\subsection{Adaptive View Fusion}
To further improve the clustering performance, we design an adaptive graph fusion method to adaptively fuse these graphs in Eq.~\eqref{xinbanben} by integrating important features from different graphs. To simplify the calculation of Eq.~\eqref{xinbanben}, borrowing the idea of alternating direction method of multipliers (ADMM)\cite{boyd2011distributed}, we can update Eq.~\eqref{xinbanben} as follows:
\begin{align}\label{adaptivenmf1}
\min\sum\limits_v&w_{v}||(M^{(v)})^{1/2}\odot U^{(v)}||_F^2+\frac{\mu}{2}||M^{(v)}||_F^2 \nonumber\\
&w_{v}=\frac{1}{2(||(M^{(v)})^{1/2}\odot U^{(v)}||_F)}\nonumber\\
&\mbox{s.t. } Y \in Def,(R^{(v)})^TR^{(v)}=I,S>0,\nonumber\\
&(M^{(v)})^TE=E,U^{(v)}=Y-F^{(v)}R^{(v)}.
\end{align}
In Eq.~\eqref{adaptivenmf1}, the difference between ($Y-F^{(v)}R^{(v)}$) and $U^{(v)}$ will affect the update of $w_{v}$, thus decrease clustering accuracy. To improve clustering results, we redefine the $w_{v}$ as follows:
\begin{align}\label{wxin}
w_{v}=\frac{1}{2(||(M^{(v)})^{1/2}\odot(Y-F^{(v)}R^{(v)})||_F)}
\end{align}
Our redefinition of $w_{v}$ is not only simple in form, but also has two advantages as follows:
\begin{enumerate}
  \item It saves the computational cost. We set the update of $Y$ and the update of $R^{(v)}$ as the first and second steps in optimization respectively (see Section~\ref{section:optimization} in more detail). Therefore, after $Y$ and $R^{(v)}$ are updated, we can update $w_{v}$ immediately with the updated $Y$ and $R^{(v)}$ (the update of $w_{v}$ can be viewed parallel with the update of other variables (i.e., $U^{(v)},M^{(v)},C^{(v)}$)). Moreover, the time complexity of updating $w_{v}$ is much smaller than the sum of time complexity of updating other variables. Therefore, for the entire optimization, the calculation of updating $w_{v}$ does not affect the total time complexity.
  \item It improves the accuracy of the calculation. It reduces the impact of difference between ($Y-F^{(v)}R^{(v)}$) and $U^{(v)}$.
\end{enumerate}

As a result, our final adaptive graph fusion method as follows:
\begin{align}\label{zuizhong}
\min\sum\limits_v&w_{v}||(M^{(v)})^{1/2}\odot U^{(v)}||_F^2+\frac{\mu}{2}||M^{(v)}||_F^2 \nonumber\\
&w_{v}=\frac{1}{2(||(M^{(v)})^{1/2}\odot(Y-F^{(v)}R^{(v)})||_F)} \nonumber\\
&\mbox{s.t. } Y \in Def,(R^{(v)})^TR^{(v)}=I,S>0,\nonumber\\
&(M^{(v)})^TE=E,U^{(v)}=Y-F^{(v)}R^{(v)}.
\end{align}

\section{Optimization}\label{section:optimization}
Since Eq.~\eqref{zuizhong} is not convex for all the variables simultaneously, inspired by ADMM~\cite{boyd2011distributed}, we leverages an iteration procedure to update these variables.

Firstly, we form the augmented Lagrangian function of Eq.~\eqref{zuizhong} as follows:
\begin{align}\label{zuizhong1}
J=\sum\limits_v&(w_{v}||(\bm{M}^{(v)})^{1/2}\odot \bm{U}^{(v)}||_F^2+\frac{\mu}{2}||\bm{M}^{(v)}||_F^2 \nonumber\\
&+\frac{\mu}{2}||\bm{Y}-\bm{F}^{(v)}\bm{R}^{(v)}-\bm{U}^{(v)}+\frac{\bm{C}^{(v)}}{\mu}||_F^2),
\end{align}
where $\bm{C}^{(v)}$ is the Lagrange multiplier of the $v$-th view.

Then, we design a six-step iteration procedure to solve Eq.~\eqref{zuizhong1}.
\begin{algorithm}[t]
    \caption{DSMC}
    \label{simc_alg}
	\begin{algorithmic}
	 \REQUIRE{Data matrix $\bm{X}^{(v)}\in \mathbb{R}^{p\times d_v}$, number of clusters $k$.}
    \STATE{Initialize weight matrix $\bm{M}^{(v)}=\bm{E}$, weight $w_v = 1/n$, $\mu=0.01, \bm{C}^{(v)}=\mathbf{0}, \mu_{max}=10^6$ for each view.}
	\REPEAT
	\STATE{Update $\bm{Y}$ by Eq. \eqref{gengy};}
	\STATE{Update $\bm{R}^{(v)}$ by Eq. \eqref{gengr};}
	\STATE{Update $\bm{M}^{(v)}$ by Eq. \eqref{gengmxin};}
	\STATE{Update $\bm{U}^{(v)}$ by Eq. \eqref{genguquan};}
	\STATE{Update $w_{v}$ by Eq. \eqref{wv};}
	\STATE{Update $\bm{C}^{(v)}$ by Eq. \eqref{cv};}
	\UNTIL{Eq. \eqref{zuizhong} converges.}
    \ENSURE{$\bm{M}^{(1)},\dots,\bm{M}^{(n)}$, $w_{1},\dots,w_{n}$ and clustering results.}
	\end{algorithmic}
\end{algorithm}

 \begin{table*}[t]
\centering
\caption{Statistics of the datasets.}
\begin{tabular}{ccccc}
\hline
Dataset  & \# of instances  &\# of features in each view & \# of views  & \# of clusters \\\hline
3 Sources       & 169& 3560 / 3631 / 3068  & 3    & 6   \\
ORL    & 400& 4096 / 3304 / 6750  & 3     & 40   \\
NUS    & 2400&64 / 114 / 73 / 128 / 225 / 500   & 6  & 12    \\
20NGs   & 500& 2000 / 2000 / 2000  & 3   & 5  \\
Scene   & 2688& 512 / 432 / 256 / 48  & 4   & 8  \\
BBC   & 685& 4659 / 4633 / 4665 / 4684  & 4   & 5  \\
\hline
\end{tabular}
\label{dataset}
\end{table*}

\textbf{Step 1.} Updating $\bm{Y}$. Fix the other variables, and the problem to solve variable $\bm{Y}$ is degraded to minimize the following problem:
\begin{align}\label{jy}
J(\bm{Y})=||\bm{Y}-\bm{F}^{(v)}\bm{R}^{(v)}-\bm{U}^{(v)}+\frac{\bm{C}^{(v)}}{\mu}||_F^2.
\end{align}

We can obtain $\bm{Y}$ by setting the derivative of $J(\bm{Y})$ w.r.t. $\bm{Y}$ to zero as follows:
\begin{align}\label{gengy}
&\frac{\partial J(\bm{Y})}{\partial \bm{Y}}=\bm{Y}-\bm{F}^{(v)}\bm{R}^{(v)}-\bm{U}^{(v)}+\frac{\bm{C}^{(v)}}{\mu}=0\nonumber\\
\Leftrightarrow&\bm{Y}=\bm{F}^{(v)}\bm{R}^{(v)}+\bm{U}^{(v)}-\frac{\bm{C}^{(v)}}{\mu}.
\end{align}
%
%

\textbf{Step 2.}
Updating $\bm{R}^{(v)}$.
Fix the other variables, and the problem to solve variable $\bm{R}^{(v)}$ is degraded to minimize the following problem:
\begin{align}\label{jr}
J(\bm{R}^{(v)})=||\bm{Y}-\bm{F}^{(v)}\bm{R}^{(v)}-\bm{U}^{(v)}+\frac{\bm{C}^{(v)}}{\mu}||_F^2.
\end{align}
Note that the objective function Eq.~\eqref{zuizhong1} w.r.t. $\bm{R}^{(v)}$ is additive and the constraints w.r.t. $\bm{R}^{(v)}$ is separable. We can update $\bm{R}^{(v)}$ individually, which is equivalent to the Orthogonal Procrustes Problem. We have following closed-form solution for it.
\begin{lemma}\label{gengr}
For problem $\min\limits_{\bm{R}^{T}\bm{R}=\bm{I}}||\bm{M}-\bm{N}\bm{R}||_F^2$, there is a closed-form solution of $\bm{R}$, that is $\bm{R}^{\ast}=\bm{UV}^T$, where $\bm{U},\bm{V}$ is constituted by the left and right singular vectors of $\bm{N}^T\bm{M}$, respectively.
\end{lemma}
Thus, for $\bm{R}^{(v)}$ in Eq.~\eqref{gengr}, it is updated according to
\begin{align}\label{gengxinr}
\bm{R}^{(v)}=\bm{U}^{(v)}\bm{V}^{(v)^T},
\end{align}
where $\bm{F}^{(v)^T}(\bm{Y}-\bm{U}^{(v)}+\frac{\bm{C}^{(v)}}{\mu})=\bm{U}^{(v)}\Sigma^{(v)}\bm{V}^{(v)^T}$.
%

\textbf{Step 3.}
Updating $\bm{M}^{(v)}$. Fix the other variables, and the problem to solve variable $\bm{M}^{(v)}$ is degraded to minimize the following problem:
\begin{align}\label{jm}
J(\bm{M}^{(v)})=&w_v||(\bm{M}^{(v)})^{1/2}\odot \bm{U}^{(v)}||_F^2 \nonumber\\
&+\frac{\mu}{2}||\bm{M}^{(v)}||_F^2.
\end{align}
When $\bm{U}^{(v)}$ is fixed, Eq.~\eqref{jm} can be rewritten as:
\begin{align}\label{gengm}
&\sum_{i=1}^{p}\sum_{j=1}^{d_v}(w_vm_{ij}^{(v)}(u_{ij}^{(v)})^2+\frac{\mu}{2}(m_{ij}^{(v)})^2)\nonumber\\
\Leftrightarrow&\sum_{i=1}^{p}\sum_{j=1}^{d_v}(m_{ij}^{(v)}+\frac{w_v(u_{ij}^{(v)})}{\mu})^2.
\end{align}
Note that for each view, Eq.~\eqref{gengm} is independent for different $j$ and we can update $\bm{M}^{(v)}$ by solving its each column $m_j^{(v)}$ separately as follows:
\begin{align}\label{gengm1}
\sum_{j=1}^{d_v}||m_j^{(v)}+\frac{1}{\mu}k_j^{(v)}||_2^2,
\end{align}
where $k_j^{(v)}$ is $j$-th column of matrix $\bm{K}^{(v)}=\bm{U}^{(v)}\odot \bm{U}^{(v)}$ for the $v$-th view.

To simplify the calculation of $m_j^{(v)}$, we reformulate Eq.~\eqref{gengm1} as the following Lagrangian function:
\begin{align}\label{gengmlage}
&L(m_j^{(v)},\alpha_j^{(v)},\beta_j^{(v)})=\frac{1}{2}||m_j^{(v)}+\frac{1}{\mu}k_j^{(v)}||_2^2\nonumber\\
&-\alpha_j^{(v)}((m_j^{(v)})^T\mathbf{1}-1)-(\beta_j^{(v)})^Tm_j^{(v)},
\end{align}
where $\alpha_j^{(v)}$ and $\beta_j^{(v)}$ are Lagrange multipliers.

Based on KKT condition, we can update $m_j^{(v)}$ as follows:
\begin{align}\label{gengm}
m_j^{(v)}=max(\alpha_j^{(v)}\mathbf{1}-\frac{1}{\mu}k_j^{(v)},0).
\end{align}
It is obvious that $(m_j^{(v)})^T\mathbf{1}=1$, and we can obtain $\alpha_j^{(v)}$ by
\begin{align}\label{gengmxin}
&\sum_{i=1}^{n}(\alpha_j^{(v)}-\frac{1}{\mu}k_j^{(v)})=1 \nonumber\\
\Leftrightarrow&\alpha_j^{(v)}=\frac{1}{p}+\frac{1}{p\mu}\sum_{i=1}^{p}k_j^{(v)}.
\end{align}
To obtain optimal $\bm{M}^{(v)}$, we can update it after calculating $\alpha_j^{(v)}$.

\textbf{Step 4.}
Updating $\bm{U}^{(v)}$. Fix the other variables, and the problem to solve variable $\bm{U}^{(v)}$ is degraded to minimize the following problem:
\begin{align}\label{ju}
&J(\bm{U}^{(v)})=w_v||(\bm{M}^{(v)})^{1/2}\odot \bm{U}^{(v)}||_F^2 \nonumber\\
&+\frac{\mu}{2}||\bm{Y}-\bm{F}^{(v)}\bm{R}^{(v)}-\bm{U}^{(v)}+\frac{\bm{C}^{(v)}}{\mu}||_F^2.
\end{align}
Define $\bm{H}^{(v)}=\bm{Y}-\bm{F}^{(v)}\bm{R}^{(v)}+\frac{\bm{C}^{(v)}}{\mu}$, and Eq.~\eqref{ju} can be rewritten as follows:
\begin{align}\label{gengu}
&w_v||(\bm{M}^{(v)})^{1/2}\odot \bm{U}^{(v)}||_F^2+\frac{\mu}{2}||\bm{U}^{(v)}-\bm{H}^{(v)}||_F^2\nonumber\\
\Leftrightarrow&\sum_{i=1}^{p}\sum_{j=1}^{d_v}(w_vm_{ij}^{(v)}(u_{ij}^{(v)})^2+\frac{\mu}{2}(m_{ij}^{(v)}-(h_{ij}^{(v)}))^2)\nonumber\\
\Leftrightarrow&\sum_{i=1}^{p}\sum_{j=1}^{d_v}(u_{ij}^{(v)}-\frac{\mu h_{ij}^{(v)}}{\mu+2w_vm_{ij}^{(v)}})^2.
\end{align}

We can obtain the optimal solution to each element $u_{ij}^{(v)}$ of variable $\bm{U}^{(v)}$ by setting Eq.~\eqref{gengu} to zero as follows:
\begin{align}\label{genguquan}
u_{ij}^{(v)}=\frac{\mu h_{ij}^{(v)}}{\mu+2w_vm_{ij}^{(v)}}.
\end{align}
\begin{table}
\centering
\caption{ACCs (\%) on six datasets.}
\begin{tabular}{ccccccc}
\hline
Method&3 Sources  & ORL & NUS  & 20NGs&Scene&BBC \\\hline
MMSC & 46.75 &26.00&13.46& 26.00&22.28 & 31.97  \\
RMSC  & 40.16& 55.17&14.40&88.69&34.57&73.25   \\
MLAN    & 68.05&72.75&11.08&92.20&51.23&44.38    \\
SwMC   & 39.64&70.75&15.92&32.6&25.19&24.53  \\
AWP   & 69.82&69.50&30.79&95.60 & 67.19&69.49  \\
DSMC   & \textbf{78.11}&\textbf{75.50}&\textbf{30.88}&\textbf{96.60}&\textbf{67.26}&\textbf{86.28} \\
\hline
\end{tabular}
\label{shiyanacc}
\end{table}
\begin{table}
\centering
\caption{NMIs (\%) on six datasets.}
\begin{tabular}{ccccccc}
\hline
Method&3 Sources  & ORL & NUS  & 20NGs&Scene&BBC \\\hline
MMSC & 30.04&48.40&2.70&3.37&7.52&5.56  \\
RMSC  & 44.55&51.59&14.26&\textbf{88.64}&33.92&67.41   \\
MLAN    & 48.04&83.84&3.04&79.85&46.94&24.76   \\
SwMC   & 11.81&83.31&6.83&16.46&15.51&4.62  \\
AWP   & 63.23&86.51&18.00&86.66&54.61&59.85  \\
DSMC   & \textbf{73.13}&\textbf{88.82}&\textbf{18.11}&87.80&\textbf{54.69}&\textbf{70.55}\\
\hline
\end{tabular}
\label{shiyannmi}
\end{table}

For Step 3 and Step 4, $\bm{M}^{(v)}$ and $\bm{U}^{(v)}$ need to be calculated iteratively based on sub-loop in theory, which will lead to excessive computation and time consumption. To simplify computation, we only update them one time in a loop.

\textbf{Step 5.}
Updating $w_{v}$. Fix the other variables, and we can update the variable $w_{v}$ by
\begin{align}\label{wv}
w_{v}=||(\bm{M}^{(v)})^{1/2}\odot (\bm{Y}-\bm{F}^{(v)}\bm{R}^{(v)})||_F.
\end{align}

\textbf{Step 6.}
Updating $\bm{C}^{(v)}$. Fix the other variables, the variable $\bm{C}^{(v)}$ can be updated by
\begin{align}\label{cv}
\bm{C}^{(v)}&=\bm{C}^{(v)}+\mu(\bm{Y}-\bm{F}^{(v)}\bm{R}^{(v)}-\bm{U}^{(v)}) \nonumber\\
\mu&=min(\mu_{max},1.1\mu),
\end{align}
where $\mu_{max}$ is the constant~\cite{wen2018incompleteb}.

The DSMC algorithm is shown in Algorithm~\ref{simc_alg}.

\section{Experiments and Analysis}
 \label{section:exp}
\subsection{Datasets}
%
%


\begin{table}[t]
\centering
\caption{Purities (\%) on six datasets.}
\begin{tabular}{ccccccc}
\hline
Method&3 Sources  & ORL & NUS  & 20NGs&Scene&BBC \\\hline
MMSC & 56.21&27.75&14.29&27.00&24.93&36.50  \\
RMSC  & 40.78&59.32&14.55&88.74&24.93&36.50   \\
MLAN    &71.60&77.25&11.33&92.20&53.65&47.30  \\
SwMC   & 44.38&76.75&16.33&33.40&26.64&25.11  \\
AWP   & 79.29&73.50&33.46&95.60&67.19&70.36  \\
DSMC   & \textbf{85.21}&\textbf{80.25}&\textbf{33.54}&\textbf{96.00}&\textbf{67.26}&\textbf{86.28}\\
\hline
\end{tabular}
\label{shiyanpurity}
\end{table}
We conduct the experiments on six real-world multi-view datasets as follows: 3 Sources\footnote{\url{http://mlg.ucd.ie/datasets/3sources.html}.}, ORL~\cite{samaria1994parameterisation}, NUS~\cite{chua2009nus}, 20 News Groups (20NGs) \footnote{\url{http://kdd.ics.uci.edu/databases/20newsgroups/20newsgroups.html}.}, Scene~\cite{fei2005bayesian} and BBC~\cite{greene2006practical}, whose important statistics summarized in Table \ref{dataset}.

\subsection{Compared Methods}
Following~\cite{nie2018multiview}, we compare our proposed method DSMC with the following state-of-the-art multi-view clustering methods:

(1)\textbf{MMSC} learns a shared graph Laplacian matrix by integrating different image features~\cite{cai2011heterogeneous}.

(2)\textbf{RMSC} learns the standard Markov chain by constructing a graph with fewer noises for each view~\cite{xia2014robust}.

(3)\textbf{MLAN} learns the local structure of multi-view data by constructing a constrained graph Laplacian~\cite{nie2017multi1}.

(4)\textbf{SwMC} learns an optimal weight for each graph by constructing a Laplacian rank constrained graph~\cite{nie2017self}.

(5)\textbf{AWP} learns optimal Procrustes for all the views adaptively~\cite{nie2018multiview}.
\begin{figure*}[t]
\centering
\subfigure[Study on 3 Sources]{\label{fig:converage_com} \includegraphics[width=0.32\textwidth]{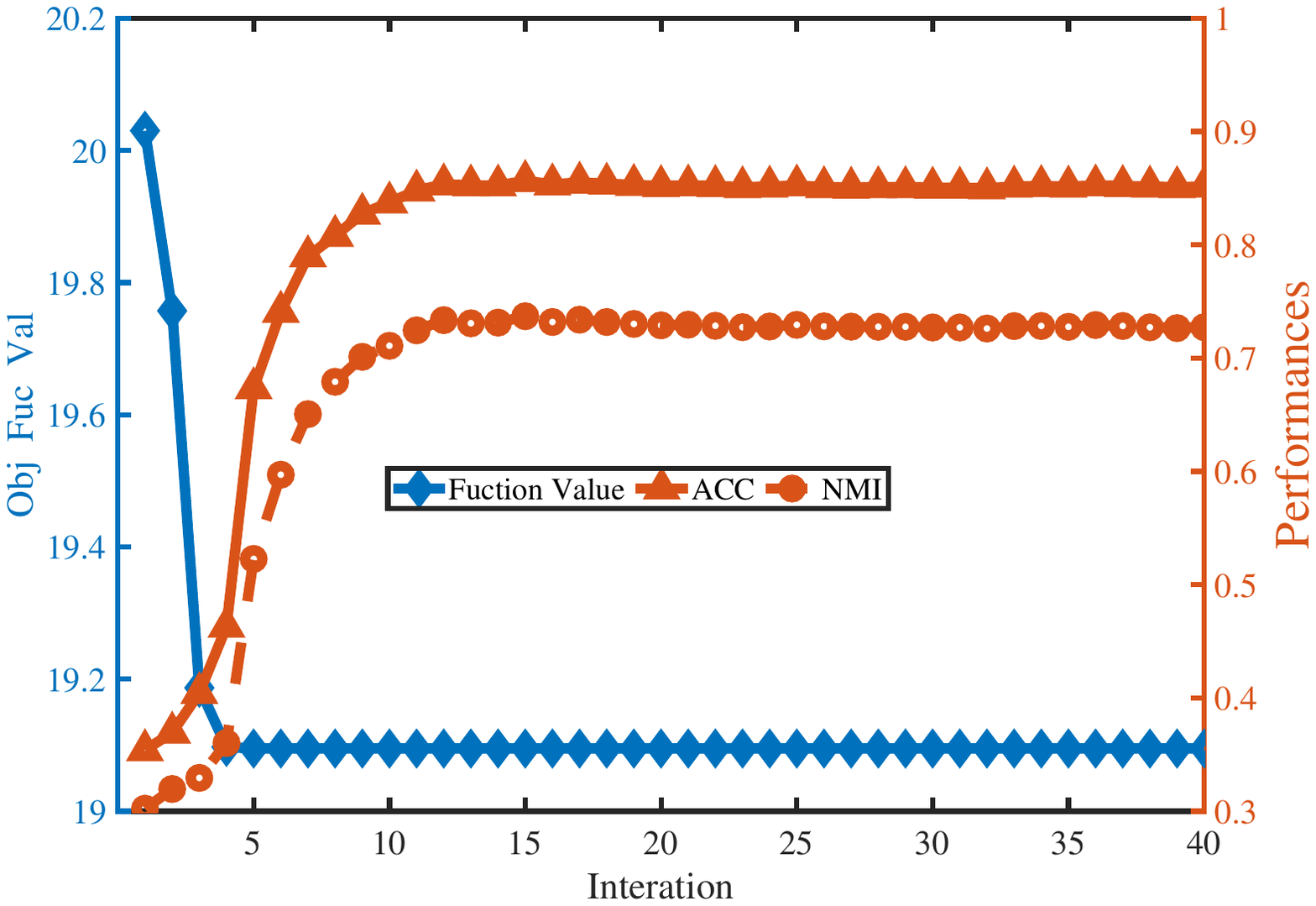} }
\subfigure[Study on ORL]{\label{fig:scene} \includegraphics[width=0.32\textwidth]{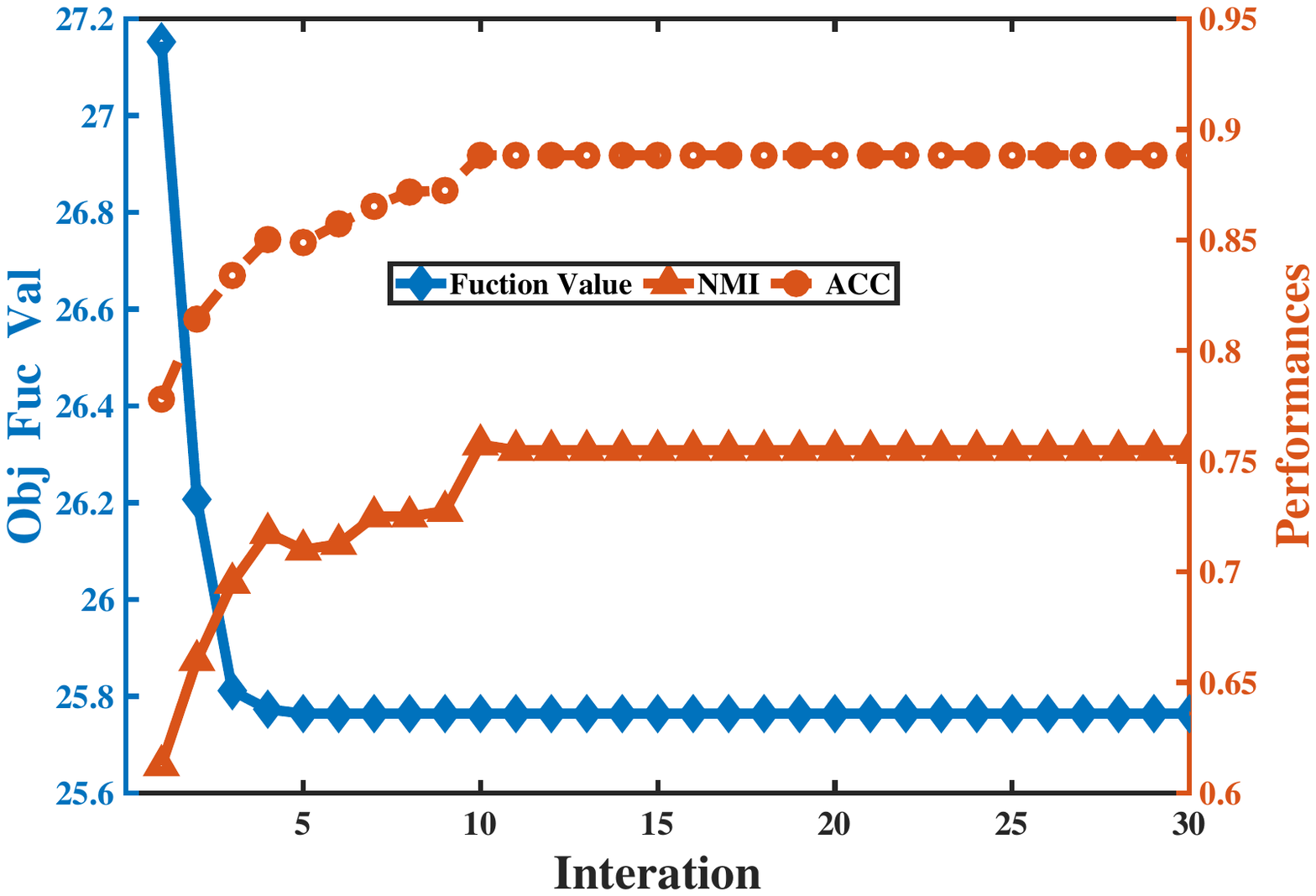}}
\subfigure[Study on NUS]{\label{fig:scene} \includegraphics[width=0.32\textwidth]{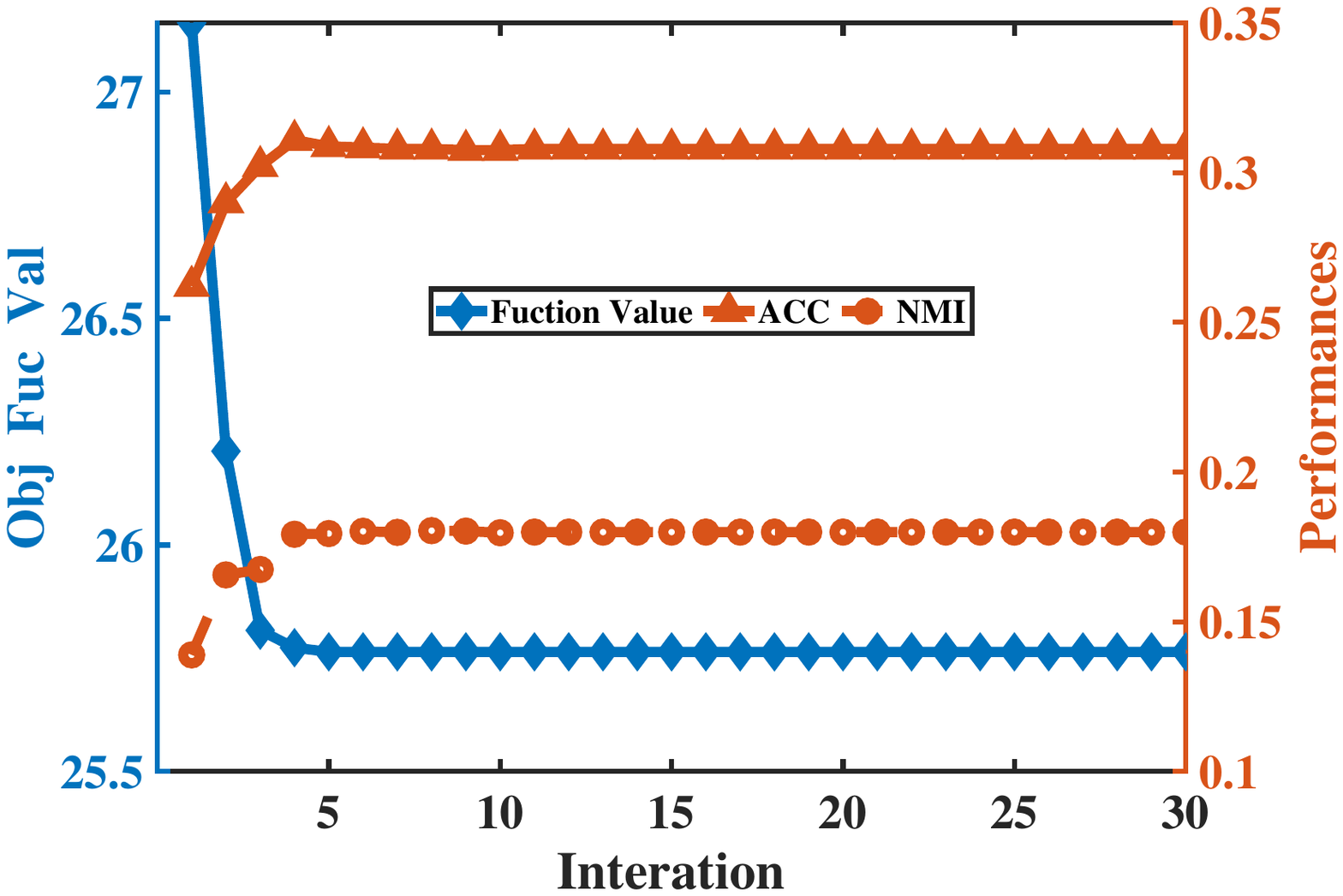}}
\subfigure[Study on 20NGs]{\label{fig:scene} \includegraphics[width=0.32\textwidth]{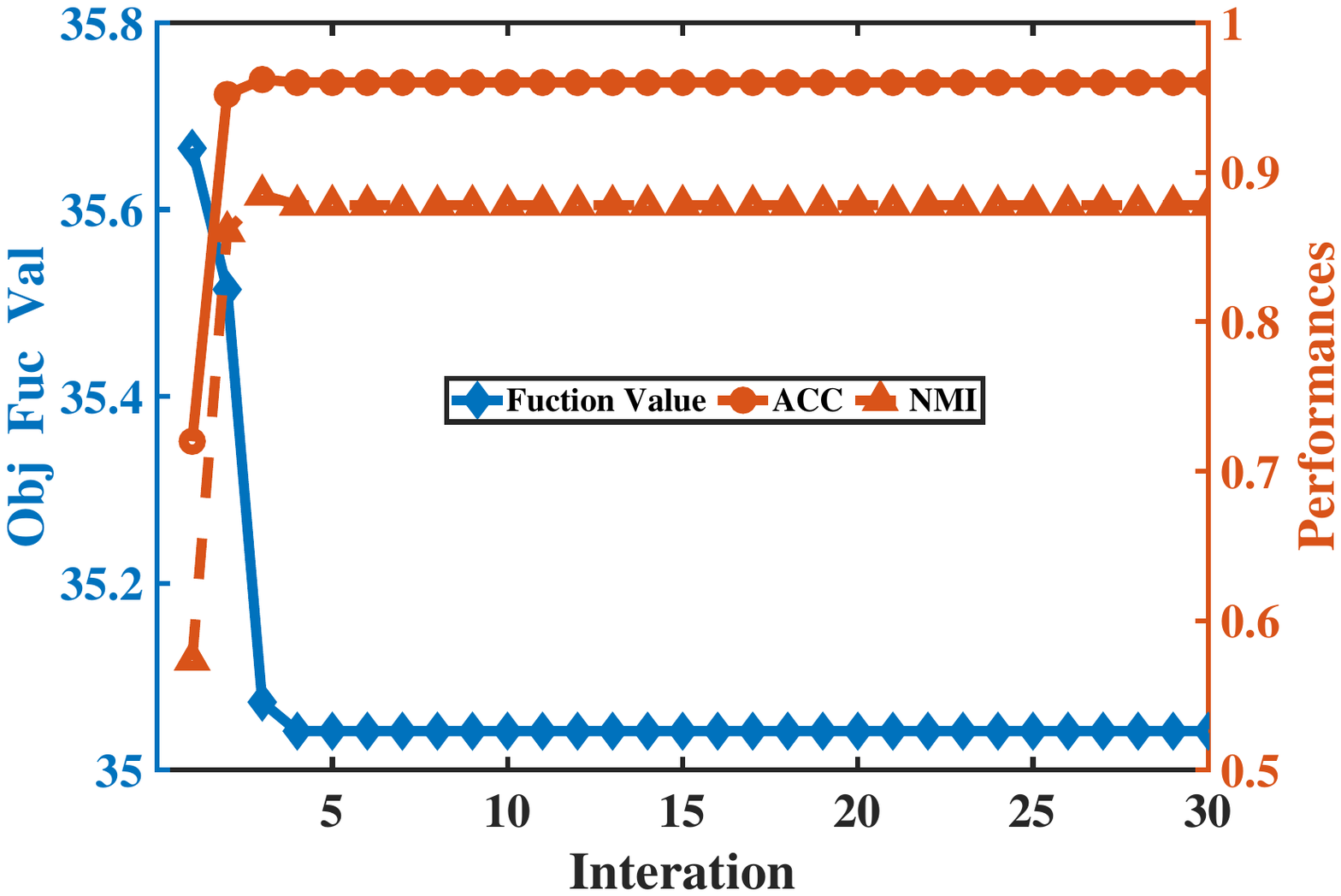}}
\subfigure[Study on Scene]{\label{fig:scene} \includegraphics[width=0.32\textwidth]{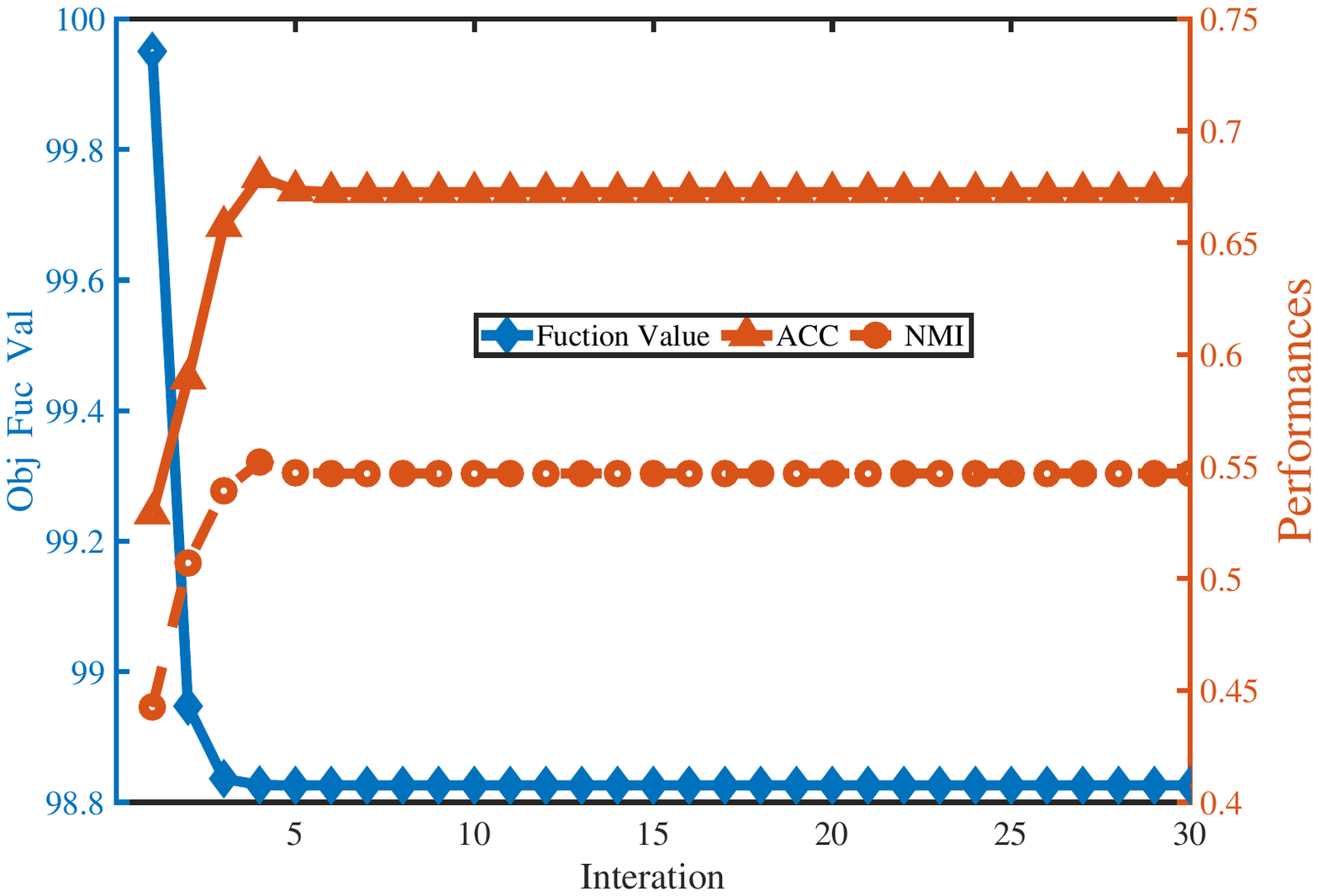}}
\subfigure[Study on BBC]{\label{fig:converage_com} \includegraphics[width=0.32\textwidth]{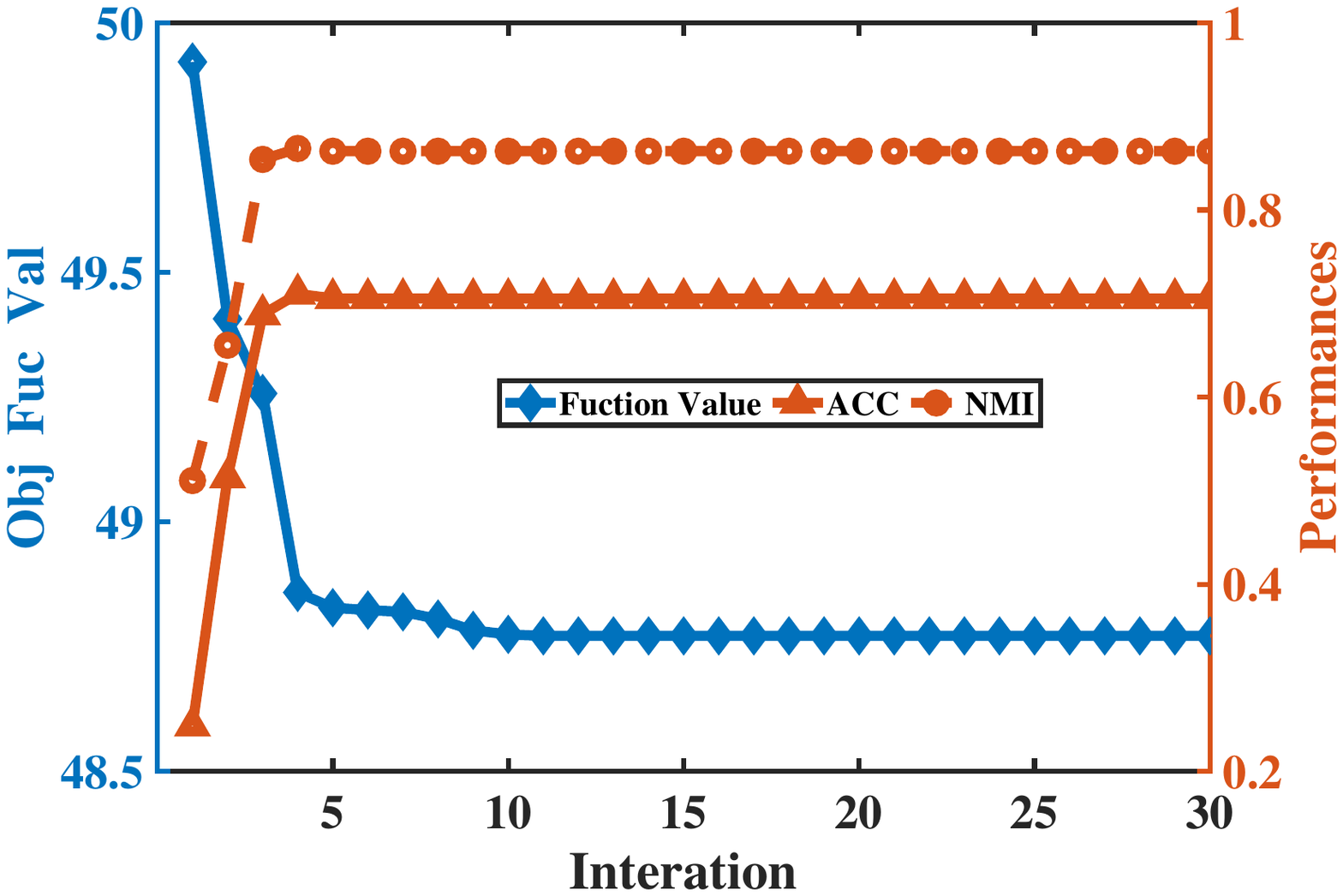} }
\caption{Convergence study on six datasets.}
\label{fig:convergency}
\end{figure*}

Following \cite{nie2018multiview}, we evaluate the results by three popular metrics: Accuracy (ACC), Normalized Mutual Information (NMI) and Purity.
\subsection{Experimental Results}
Table \ref{shiyanacc} to Table \ref{shiyanpurity} shows the ACC, NMI and Purity values on six real-world datasets, respectively. Bold numbers denote the best result.

\paragraph{Results on 3 Sources dataset:}
DSMC outperforms all the other methods significantly on 3 Sources. Specifically, compared with SwMC, we improve ACC by $38.47\%$, NMI by $61.32\%$, Purity by $40.83\%$. Relative to the latest method AWP, DSMC improves ACC by $8.29\%$, NMI by $9.90\%$, Purity by $5.92\%$. One possible reason for our outstanding clustering performance is that each view of 3 Sources dataset has a large number of important features, and our methods can effectively extract these important features.


\paragraph{Results on ORL dataset:}
DSMC has better performances than the other methods on ORL. Obviously, compared with MMSC, DSMC improves ACC by $49.5\%$, NMI by $40.42\%$, Purity by $52.5\%$. Relative to AWP, it improves ACC by $6.00\%$, NMI by $2.31\%$, Purity by $6.75\%$. It is because each view of ORL has a large number of features, DSMC can distinguish the importance of these features by adaptively weighting them.

\paragraph{Results on NUS dataset:}
DSMC outperforms all the other methods significantly on NUS. Especially, compared with MLAN, our DSMC improves ACC by $19.8\%$, NMI by $15.07\%$, Purity by $22.21\%$. Relative to the latest method AWP, DSMC improves ACC by $8.29\%$, NMI by $9.90\%$, Purity by $5.92\%$. The main reason is that NUS has $6$ views and our DSMC can effectively integrate favorable features from different graphs.

\paragraph{Results on 20NGs dataset:}
For 20NGs dataset, DSMC obtains better clustering results than most of the other methods. Moreover, compared with MMSC, it improves ACC by $70.6\%$, NMI by $84.43\%$, Purity by $69\%$. Relative to the latest method AWP, DSMC improves ACC by $8.29\%$, NMI by $9.90\%$, Purity by $5.92\%$. The main reason for this phenomenon is that 20NGs dataset has many unfavorable features, which plays a negative role in learning optimal graphs. But our DSMC is able to remove these unfavorable features based on the weight matrix $M^{(v)}$. RMSC also achieves satisfactory clustering results. It is because each view of 20NGs has the same feature dimension (2000 features), which helps RMSC obtain the satisfactory Markov chain.

\paragraph{Results on Scene dataset:}
Obviously, our proposed DSMC outperforms all the other methods significantly on Scene. In addition, compared with MMSC, DSMC improves ACC by $44.98\%$, NMI by $47.17\%$, Purity by $42.33\%$. AWP can achieve good clustering results because each view of this dataset contains a small number of features, which makes it easy for AWP to learn optimal Procrustes. Compared with AWP, our proposed DSMC improves ACC by $7\%$, NMI by $8\%$, Purity by $7\%$. This is mainly because DSMC is able to weigh different features through weight matrix $\bm{M}^{(v)}$, thereby extracting favorable features from views with different feature dimensions.

\paragraph{Results on BBC dataset:}
When clustering BBC dataset, DSMC has better clustering performance than the other methods. Specifically, compared with SwMC, our proposed DSMC raises ACC by $61.75\%$, NMI by $65.93\%$, Purity by $62.17\%$. Relative to the latest method AWP, DSMC improves ACC by $16.79\%$, NMI by $10.7\%$, Purity by $15.92\%$. The main reason is that our DSMC can adaptively integrate different features multiple graphs fusion.

In summary, MMSC and SwMC often obtain poor clustering results. It is because they only learn one weight for different views, and do not take into account the local information on different features in each view. When clustering these datasets with a large number of instances (e.g., Scene and NUS), these local information will have a great impact on our clustering results. Meanwhile, although AWP can extract local information of the data, it cannot distinguish the importance of different features. Therefore, when we cluster multi-view data with high-dimensional features (e.g., BBC), AWP cannot obtain satisfactory results. By weighting different features, our proposed DSMC can effectively integrate different features, and adaptively fuse different graphs. Finally, DSMC always achieves the best performances among the compared state-of-the-art multi-view clustering methods.
%


\subsection{Convergence Study}
In Figure~\ref{fig:convergency}, we show the convergence curve, ACC values and NMI values w.r.t. the number of iterations\footnote{``Obj Fun Val" is the abbreviation of ``Objetive Function Value".}. The blue solid curve indicates the value of the objective function and the dashed lines represent the clustering performance (ACC and NMI) of our proposed DSMC.

Obviously, for all the datasets, DSMC converges in less than $10$ steps, which proves the effectiveness and efficiency of our adaptive multiple graphs fusion. It is because after each iteration, DSMC can extract the favorable features of each view with the help of weight matrix $M^{(v)}$, and integrate different features of each view by $w_v$. When suitable weights are learned, the gradient of the objective function (Eq.~\eqref{zuizhong}) is close to the ideal gradient, which accelerates the convergence of the objective function.

\section{Conclusion}  \label{section:con}
In this paper, we propose an effective dual self-weighted multi-view clustering framework, named DSMC. DSMC can remove redundant features and noises from each graph to improve clustering performance. In the framework of DSMC, we assign different weights to different features by imposing an adaptive weight factor. We design an adaptive multiple graphs fusion to fuse the features in the different views, thus integrating different graphs for clustering. We perform experiments on six real-world multi-view datasets to show the effectiveness and efficiency of our DSMC.

%
%
%
%
%

\ifCLASSOPTIONcaptionsoff
  \newpage
\fi
\bibliographystyle{IEEEtran}
\bibliography{IEEEtran}

@InProceedings{lu2013unified,
  author    = {Lu, Zhiwu and Peng, Yuxin},
  title     = {Unified constraint propagation on multi-view data},
  booktitle = {Twenty-Seventh AAAI Conference on Artificial Intelligence},
  year      = {2013},
}

@InProceedings{nie2017self,
  author    = {Nie, Feiping and Li, Jing and Li, Xuelong and others},
  title     = {Self-weighted Multiview Clustering with Multiple Graphs.},
  booktitle = {IJCAI},
  year      = {2017},
  pages     = {2564--2570},
}

@InProceedings{kang2018self,
  author       = {Kang, Zhao and Lu, Xiao and Yi, Jinfeng and Xu, Zenglin},
  title        = {Self-weighted multiple kernel learning for graph-based clustering and semi-supervised classification},
  booktitle    = {Proceedings of the 27th International Joint Conference on Artificial Intelligence},
  year         = {2018},
  pages        = {2312--2318},
  organization = {AAAI Press},
}

@Article{li2018multi,
  author    = {Li, Sheng and Shao, Ming and Fu, Yun},
  title     = {Multi-view low-rank analysis with applications to outlier detection},
  journal   = {ACM Transactions on Knowledge Discovery from Data (TKDD)},
  year      = {2018},
  volume    = {12},
  number    = {3},
  pages     = {32},
  publisher = {ACM},
}

@InProceedings{du2018multi,
  author       = {Du, Changying and Du, Changde and Xie, Xingyu and Zhang, Chen and Wang, Hao},
  title        = {Multi-view Adversarially Learned Inference for Cross-domain Joint Distribution Matching},
  booktitle    = {Proceedings of the 24th ACM SIGKDD International Conference on Knowledge Discovery \& Data Mining},
  year         = {2018},
  pages        = {1348--1357},
  organization = {ACM},
}

@Article{gong2019multi,
  author    = {Gong, Chen and Yang, Jian and Tao, Dacheng},
  title     = {Multi-Modal Curriculum Learning over Graphs},
  journal   = {ACM Transactions on Intelligent Systems and Technology (TIST)},
  year      = {2019},
  volume    = {10},
  number    = {4},
  pages     = {35},
  publisher = {ACM},
}

@InProceedings{cai2011heterogeneous,
  author       = {Cai, Xiao and Nie, Feiping and Huang, Heng and Kamangar, Farhad},
  title        = {Heterogeneous image feature integration via multi-modal spectral clustering},
  booktitle    = {CVPR 2011},
  year         = {2011},
  pages        = {1977--1984},
  organization = {IEEE},
}

@InProceedings{Tang2019AAAI,
  author    = {Tang, Chang and Zhu, Xinzhong and Liu, Xinwang and Wang, Lizhe},
  title     = {Cross-view Local Structure Preserved Diversity and Consensus Learning for Multi-view Unsupervised Feature Selection},
  booktitle = {AAAI Conference on Artificial Intelligence},
  year      = {2019},
  //pages   = {595--604},
}

@InProceedings{blum1998combining,
  author       = {Blum, Avrim and Mitchell, Tom},
  title        = {Combining labeled and unlabeled data with co-training},
  booktitle    = {Proceedings of the eleventh annual conference on Computational learning theory},
  year         = {1998},
  pages        = {92--100},
  organization = {ACM},
}

@InProceedings{fei2005bayesian,
  author       = {Fei-Fei, Li and Perona, Pietro},
  title        = {A bayesian hierarchical model for learning natural scene categories},
  booktitle    = {2005 IEEE Computer Society Conference on Computer Vision and Pattern Recognition (CVPR'05)},
  year         = {2005},
  volume       = {2},
  pages        = {524--531},
  organization = {IEEE},
}

@InProceedings{lan2015reducing,
  author       = {Lan, Chao and Huan, Jun},
  title        = {Reducing the unlabeled sample complexity of semi-supervised multi-view learning},
  booktitle    = {Proceedings of the 21th ACM SIGKDD International Conference on Knowledge Discovery and Data Mining},
  year         = {2015},
  pages        = {627--634},
  organization = {ACM},
}

@Article{wen2018incompleteb,
  author    = {Wen, Jie and Xu, Yong and Liu, Hong},
  title     = {Incomplete Multiview Spectral Clustering With Adaptive Graph Learning},
  journal   = {IEEE transactions on cybernetics},
  year      = {2018},
  publisher = {IEEE},
}

@Article{boyd2011distributed,
  author    = {Boyd, Stephen and Parikh, Neal and Chu, Eric and Peleato, Borja and Eckstein, Jonathan and others},
  title     = {Distributed optimization and statistical learning via the alternating direction method of multipliers},
  journal   = {Foundations and Trends{\textregistered} in Machine learning},
  year      = {2011},
  volume    = {3},
  number    = {1},
  pages     = {1--122},
  publisher = {Now Publishers, Inc.},
}

@InProceedings{xia2014robust,
  author    = {Xia, Rongkai and Pan, Yan and Du, Lei and Yin, Jian},
  title     = {Robust multi-view spectral clustering via low-rank and sparse decomposition},
  booktitle = {Twenty-Eighth AAAI Conference on Artificial Intelligence},
  year      = {2014},
}

@InProceedings{greene2006practical,
  author       = {Greene, Derek and Cunningham, P{\'a}draig},
  title        = {Practical solutions to the problem of diagonal dominance in kernel document clustering},
  booktitle    = {Proceedings of the 23rd international conference on Machine learning},
  year         = {2006},
  pages        = {377--384},
  organization = {ACM},
}

@InProceedings{nie2018multiview,
  author       = {Nie, Feiping and Tian, Lai and Li, Xuelong},
  title        = {Multiview clustering via adaptively weighted procrustes},
  booktitle    = {Proceedings of the 24th ACM SIGKDD International Conference on Knowledge Discovery \& Data Mining},
  year         = {2018},
  pages        = {2022--2030},
  organization = {ACM},
}

@InProceedings{nie2017multi1,
  author    = {Nie, Feiping and Cai, Guohao and Li, Xuelong},
  title     = {Multi-view clustering and semi-supervised classification with adaptive neighbours},
  booktitle = {AAAI},
  year      = {2017},
}

@InProceedings{samaria1994parameterisation,
  author       = {Samaria, Ferdinando S and Harter, Andy C},
  title        = {Parameterisation of a stochastic model for human face identification},
  booktitle    = {Proceedings of 1994 IEEE workshop on applications of computer vision},
  year         = {1994},
  pages        = {138--142},
  organization = {IEEE},
}

@InProceedings{chua2009nus,
  author    = {Chua, Tat-Seng and Tang, Jinhui and Hong, Richang and Li, Haojie and Luo, Zhiping and Zheng, Yantao},
  title     = {NUS-WIDE: a real-world web image database from National University of Singapore},
  booktitle = {Proceedings of the ACM international conference on image and video retrieval},
  year      = {2009},
  pages     = {1--9},
}

@InProceedings{min2017delicious,
  author    = {Min, Weiqing and Jiang, Shuqiang and Wang, Shuhui and Sang, Jitao and Mei, Shuhuan},
  title     = {A delicious recipe analysis framework for exploring multi-modal recipes with various attributes},
  booktitle = {ACM MM},
  year      = {2017},
  pages     = {402--410},
}

@InProceedings{wang2013collaborative,
  author    = {Wang, Hao and Chen, Binyi and Li, Wu-Jun},
  title     = {Collaborative topic regression with social regularization for tag recommendation},
  booktitle = {IJCAI},
  year      = {2013},
}

@inproceedings{wang2019compact,
  title={A compact and language-sensitive multilingual translation method},
  author={Wang, Yining and Zhou, Long and Zhang, Jiajun and Zhai, Feifei and Xu, Jingfang and Zong, Chengqing},
  booktitle={Proceedings of the 57th Annual Meeting of the Association for Computational Linguistics},
  pages={1213--1223},
  year={2019}
}

@Book{friedman2001elements,
  title     = {The elements of statistical learning},
  publisher = {Springer series in statistics New York},
  year      = {2001},
  author    = {Friedman, Jerome and Hastie, Trevor and Tibshirani, Robert},
  volume    = {1},
  number    = {10},
}

@article{fang2025your,
  title={Your data is not perfect: Towards cross-domain out-of-distribution detection in class-imbalanced data},
  author={Fang, Xiang and Easwaran, Arvind and Genest, Blaise and Suganthan, Ponnuthurai Nagaratnam},
  journal={Expert Systems with Applications},
  year={2025}
}

@article{fang2023hierarchical,
  title={Hierarchical local-global transformer for temporal sentence grounding},
  author={Fang, Xiang and Liu, Daizong and Zhou, Pan and Xu, Zichuan and Li, Ruixuan},
  journal={IEEE Transactions on Multimedia},
  year={2023},
  publisher={IEEE}
}

@article{fang2022multi,
  title={Multi-modal cross-domain alignment network for video moment retrieval},
  author={Fang, Xiang and Liu, Daizong and Zhou, Pan and Hu, Yuchong},
  journal={IEEE Transactions on Multimedia},
  volume={25},
  pages={7517--7532},
  year={2022},
  publisher={IEEE}
}

@inproceedings{fang2026cogniVerse,
  title={CogniVerse: Revolutionizing Multi-modal Retrieval-Augmented Generation with Cognitive Reflection and Geometric Reasoning},
  author={Fang, Xiang and Fang, Wanlong and Wang, Changshuo},
  booktitle={Proceedings of the IEEE/CVF Conference on Computer Vision and Pattern Recognition},
  year={2026}
}

@inproceedings{fang2023you,
  title={You can ground earlier than see: An effective and efficient pipeline for temporal sentence grounding in compressed videos},
  author={Fang, Xiang and Liu, Daizong and Zhou, Pan and Nan, Guoshun},
  booktitle={Proceedings of the IEEE/CVF Conference on Computer Vision and Pattern Recognition},
  pages={2448--2460},
  year={2023}
}

@inproceedings{fang2025hierarchical,
  title={Hierarchical Semantic-Augmented Navigation: Optimal Transport and Graph-Driven Reasoning for Vision-Language Navigation},
  author={Fang, Xiang and Fang, Wanlong and Wang, Changshuo},
  booktitle={Advances in Neural Information Processing Systems},
  year={2025}
}

@inproceedings{fang2025adaptive,
  title={Adaptive Multi-prompt Contrastive Network for Few-shot Out-of-distribution Detection},
  author={Fang, Xiang and Easwaran, Arvind and Genest, Blaise},
  booktitle={International Conference on Machine Learning},
  year={2025}
}

@inproceedings{fang2026slap,
  title={SLAP: The Semantic Least Action Principle for Variational Video-Language Modeling},
  author={Fang, Xiang and Fang, Wanlong},
  booktitle={International Conference on Machine Learning},
  year={2026}
}

@inproceedings{fang2026immuno,
  title={Immuno-VLM: Immunizing Large Vision-Language Models via Generative Semantic Antibodies for Open-World Trustworthiness},
  author={Fang, Xiang and Fang, Wanlong and Ji, Wei},
  booktitle={International Conference on Machine Learning},
  year={2026}
}

@inproceedings{fang2026disentangling,
  title={Disentangling Adversarial Prompts: A Semantic-Graph Defense for Robust LLM Security},
  author={Fang, Xiang and Fang, Wanlong},
 booktitle={Proceedings of the AAAI Conference on Artificial Intelligence},
year={2026}
}

@inproceedings{fang2026advancing,
  title={Advancing Out-of-Distribution Detection Across Diverse Scenarios},
  author={Fang, Xiang},
  booktitle={Proceedings of the AAAI Conference on Artificial Intelligence},
  volume={40},
  number={48},
  pages={41042--41043},
  year={2026}
}

@inproceedings{fang2026unveiling,
  title={Unveiling the Fragility of Vision-Language Models: Multi-Modal Adversarial Synergy via Texture-Constrained Perturbations and Cross-Modal Optimization},
  author={Fang, Xiang and Fang, Wanlong and Wang, Changshuo},
 booktitle={Proceedings of the AAAI Conference on Artificial Intelligence},
year={2026}
}

@inproceedings{fang2026rethinking,
  title={Rethinking Video-language Model From the Language Input Perspective},
  author={Fang, Xiang and Fang, Wanlong and Wang, Changshuo and Qu, Xiaoye and Liu, Daizong},
 booktitle={Proceedings of the AAAI Conference on Artificial Intelligence},
year={2026}
}

@inproceedings{fang2026towards,
  title={Towards Unified Vision-Language Models With Incomplete Multi-Modal Inputs},
  author={Fang, Xiang and Fang, Wanlong and Wang, Changshuo and Tang, Keke and Liu, Daizong and Wang, Siyi and Ji, Wei},
 booktitle={Proceedings of the AAAI Conference on Artificial Intelligence},
year={2026}
}

@inproceedings{fang2025multi,
  title={Multi-pair temporal sentence grounding via multi-thread knowledge transfer network},
  author={Fang, Xiang and Fang, Wanlong and Wang, Changshuo and Liu, Daizong and Tang, Keke and Dong, Jianfeng and Zhou, Pan and Li, Beibei},
  booktitle={Proceedings of the AAAI Conference on Artificial Intelligence},
  volume={39},
  number={3},
  pages={2915--2923},
  year={2025}
}

@inproceedings{fang2024fewer,
  title={Fewer Steps, Better Performance: Efficient Cross-Modal Clip Trimming for Video Moment Retrieval Using Language},
  author={Fang, Xiang and Liu, Daizong and Fang, Wanlong and Zhou, Pan and Xu, Zichuan and Xu, Wenzheng and Chen, Junyang and Li, Renfu},
  booktitle={Proceedings of the AAAI Conference on Artificial Intelligence},
  volume={38},
  number={2},
  pages={1735--1743},
  year={2024}
}

@inproceedings{fang2024multi,
  title={Multi-Pair Temporal Sentence Grounding via Multi-Thread Knowledge Transfer Network},
  author={Fang, Xiang and Fang, Wanlong and Wang, Changshuo and Liu, Daizong and Tang, Keke and Dong, Jianfeng and Zhou, Pan and Li, Beibei},
  booktitle={Proceedings of the AAAI Conference on Artificial Intelligence},
  year={2025}
}

@inproceedings{fang2025turing,
  title={Turing Patterns for Multimedia: Reaction-Diffusion Multi-Modal Fusion for Language-Guided Video Moment Retrieval},
  author={Fang, Xiang and Fang, Wanlong and Ji, Wei and Chua, Tat-Seng},
  booktitle={ACM International Conference on Multimedia},
  year={2025}
}

@inproceedings{fang2024not,
  title={Not all inputs are valid: Towards open-set video moment retrieval using language},
  author={Fang, Xiang and Fang, Wanlong and Liu, Daizong and Qu, Xiaoye and Dong, Jianfeng and Zhou, Pan and Li, Renfu and Xu, Zichuan and Chen, Lixing and Zheng, Panpan and others},
  booktitle={Proceedings of the 32nd ACM International Conference on Multimedia},
  pages={28--37},
  year={2024}
}

@inproceedings{fang2024rethinking,
  title={Rethinking Weakly-supervised Video Temporal Grounding From a Game Perspective},
  author={Fang, Xiang and Xiong, Zeyu and Fang, Wanlong and Qu, Xiaoye and Chen, Chen and Dong, Jianfeng and Tang, Keke and Zhou, Pan and Cheng, Yu and Liu, Daizong},
  booktitle={European Conference on Computer Vision},
  year={2024},
  organization={Springer}
}

@inproceedings{fang2023annotations,
  title={Annotations Are Not All You Need: A Cross-modal Knowledge Transfer Network for Unsupervised Temporal Sentence Grounding},
  author={Fang, Xiang and Liu, Daizong and Fang, Wanlong and Zhou, Pan and Cheng, Yu and Tang, Keke and Zou, Kai},
  booktitle={Findings of the Association for Computational Linguistics: EMNLP 2023},
  pages={8721--8733},
  year={2023}
}

@article{fang2021unbalanced,
  title={Unbalanced incomplete multi-view clustering via the scheme of view evolution: Weak views are meat; strong views do eat},
  author={Fang, Xiang and Hu, Yuchong and Zhou, Pan and Wu, Dapeng Oliver},
  journal={IEEE Transactions on Emerging Topics in Computational Intelligence},
  volume={6},
  number={4},
  pages={913--927},
  year={2021},
  publisher={IEEE}
}

@article{fang2025adaptivetai,
  title={Adaptive Hierarchical Graph Cut for Multi-granularity Out-of-distribution Detection},
  author={Fang, Xiang and Easwaran, Arvind and Genest, Blaise and Suganthan, Ponnuthurai Nagaratnam},
  journal={IEEE Transactions on Artificial Intelligence},
  year={2025}
}

@article{fang2021animc,
  title={Animc: A soft approach for autoweighted noisy and incomplete multiview clustering},
  author={Fang, Xiang and Hu, Yuchong and Zhou, Pan and Wu, Dapeng},
  journal={IEEE Transactions on Artificial Intelligence},
  volume={3},
  number={2},
  pages={192--206},
  year={2021},
  publisher={IEEE}
}

@article{fang2020v,
  title={V3H: View variation and view heredity for incomplete multiview clustering},
  author={Fang, Xiang and Hu, Yuchong and Zhou, Pan and Wu, Dapeng Oliver},
  journal={IEEE Transactions on Artificial Intelligence},
  volume={1},
  number={3},
  pages={233--247},
  year={2020},
  publisher={IEEE}
}

@article{liu2023exploring,
  title={Exploring optical-flow-guided motion and detection-based appearance for temporal sentence grounding},
  author={Liu, Daizong and Fang, Xiang and Hu, Wei and Zhou, Pan},
  journal={IEEE Transactions on Multimedia},
  volume={25},
  pages={8539--8553},
  year={2023},
  publisher={IEEE}
}

@inproceedings{wang2025taylor,
  title={Taylor series-inspired local structure fitting network for few-shot point cloud semantic segmentation},
  author={Wang, Changshuo and He, Shuting and Fang, Xiang and Wu, Meiqing and Lam, Siew-Kei and Tiwari, Prayag},
  booktitle={Proceedings of the AAAI Conference on Artificial Intelligence},
  volume={39},
  number={7},
  pages={7527--7535},
  year={2025}
}

@inproceedings{wang2025point,
  title={Point clouds meets physics: Dynamic acoustic field fitting network for point cloud understanding},
  author={Wang, Changshuo and He, Shuting and Fang, Xiang and Han, Jiawei and Liu, Zhonghang and Ning, Xin and Li, Weijun and Tiwari, Prayag},
  booktitle={Proceedings of the Computer Vision and Pattern Recognition Conference},
  pages={22182--22192},
  year={2025}
}

@inproceedings{wang2025dypolyseg,
  title={DyPolySeg: Taylor Series-Inspired Dynamic Polynomial Fitting Network for Few-shot Point Cloud Semantic Segmentation},
  author={Wang, Changshuo and Fang, Xiang and Tiwari, Prayag},
  booktitle={Forty-second International Conference on Machine Learning},
  year={2025}
}

@article{wang2026reasoning,
  title={Reasoning beyond points: A visual introspective approach for few-shot 3d segmentation},
  author={Wang, Changshuo and He, Shuting and Fang, Xiang and Hu, Zhijian and Huang, Jia-Hong and Shen, Yixian and Tiwari, Prayag},
  journal={Advances in Neural Information Processing Systems},
  volume={38},
  pages={117394--117414},
  year={2026}
}

@article{wang2026from,
  title={From Coarse to Fine: Deep Prototype Refinement Network for Few-Shot Point Cloud Semantic Segmentation},
  author={Wang, Changshuo and He, Shuting and Fang, Xiang and Li, Weijun and Gao, Xingyu and Liu, Zhonghang and Tiwari, Prayag and Kanoulas, Dimitrios},
  journal={International Conference on Machine Learning},
  year={2026}
}

@article{wang2026topadapter,
  title={TopAdapter: Topology-Aware Prompt Tuning for Efficient Point Cloud Understanding},
  author={Wang, Changshuo and He, Shuting and Fang, Xiang and Li, Weijun and Shen, Yixian and Xu, Mingkun and Sun, Zhongtian and Tiwari, Prayag},
  journal={International Conference on Machine Learning},
  year={2026}
}

@inproceedings{wang2026biologically,
  title={Biologically-Inspired Evolutionary Domain Symbiosis for Few-shot and Zero-shot Point Cloud Semantic Segmentation},
  author={Wang, Changshuo and Hu, Zhijian and Fang, Xiang and Yu, Zai Yang and Wu, Yibin and Xu, Mingkun and Wang, Yusong and Gao, Xingyu and Tiwari, Prayag},
  booktitle={Proceedings of the AAAI Conference on Artificial Intelligence},
  volume={40},
  number={12},
  pages={9666--9674},
  year={2026}
}

@inproceedings{yang2025eood,
  title={EOOD: Entropy-based Out-of-distribution Detection},
  author={Yang, Guide and Hou, Chao and Peng, Weilong and Fang, Xiang and Nie, Yongwei and Zhu, Peican and Tang, Keke},
  booktitle={2025 International Joint Conference on Neural Networks (IJCNN)},
  pages={1--8},
  year={2025},
  organization={IEEE}
}

@inproceedings{wang2025reducing
,
  title={Reducing T-Depth and T-Count in Quantum Multiplication Using Compressor Primitives},
  author={Wang, Siyi and Dutta, Suman and Lee, Wei Jie Bryan and Feng, Jerrie and Fang, Xiang and Chattopadhyay, Anupam},
  booktitle={Proceedings of the Great Lakes Symposium on VLSI 2025},
  pages={35--40},
  year={2025}
}

@inproceedings{lei2025exploring,
  title={Exploring Disentangled Appearance-Motion Contexts for Temporal Activity Localization},
  author={Lei, Huashuo and Cai, Xiaowen and Liu, Daizong and Fang, Xiang and Qu, Xiaoye and Dong, Jianfeng and Yu, Jixiang and Jin, Keyan},
  booktitle={2025 International Joint Conference on Neural Networks (IJCNN)},
  pages={1--8},
  year={2025},
  organization={IEEE}
}

@inproceedings{zhang2025monoattack,
  title={MonoAttack: A Strong Attack Framework with Depth-Migration and Attribute-Tampering for Monocular 3D Object Detection},
  author={Zhang, Xiayue and Lei, Huashuo and Liu, Daizong and Qu, Xiaoye and Fang, Xiang and Guan, Runwei and Jin, Keyan},
  booktitle={2025 International Joint Conference on Neural Networks (IJCNN)},
  pages={1--8},
  year={2025},
  organization={IEEE}
}

@inproceedings{zhang2025manipulating,
  title={Manipulating the Bounding Box: Multimodal Controlled Backdoor Attacks on 3D Visual Grounding Models},
  author={Zhang, Xiayue and Lei, Huashuo and Liu, Daizong and Qu, Xiaoye and Fang, Xiang and Guan, Runwei and Jin, Keyan},
  booktitle={2025 International Joint Conference on Neural Networks (IJCNN)},
  pages={1--8},
  year={2025},
  organization={IEEE}
}

@article{wang2025prototype,
  title={Prototype-driven structure synergy network for remote sensing images segmentation},
  author={Wang, Junyi and Li, Jinjiang and Fan, Guodong and Ju, Yakun and Fang, Xiang and Kot, Alex C},
  journal={IEEE Transactions on Geoscience and Remote Sensing},
  year={2025},
  publisher={IEEE}
}

@inproceedings{wang2025seeing,
  title={Seeing the Overlooked: Bio-Visual Inspired Weak Saliency Feedback Transformer for Person Re-identification},
  author={Wang, Changshuo and He, Shuting and Fang, Xiang and Nan, Fangzhe and Tiwari, Prayag},
  booktitle={Proceedings of the 33rd ACM International Conference on Multimedia},
  pages={3192--3201},
  year={2025}
}

@inproceedings{fang2026align,
  title={To align or not to align: Strategic multimodal representation alignment for optimal performance},
  author={Fang, Wanlong and Zhang, Tianle and Chan, Alvin},
  booktitle={Proceedings of the AAAI Conference on Artificial Intelligence},
  volume={40},
  number={25},
  pages={21056--21064},
  year={2026}
}

@article{liu2023conditional,
  title={Conditional video diffusion network for fine-grained temporal sentence grounding},
  author={Liu, Daizong and Zhu, Jiahao and Fang, Xiang and Xiong, Zeyu and Wang, Huan and Li, Renfu and Zhou, Pan},
  journal={IEEE Transactions on Multimedia},
  volume={26},
  pages={5461--5476},
  year={2023},
  publisher={IEEE}
}

@article{liu2024pandora,
  title={Pandora's box: Towards building universal attackers against real-world large vision-language models},
  author={Liu, Daizong and Yang, Mingyu and Qu, Xiaoye and Zhou, Pan and Fang, Xiang and Tang, Keke and Wan, Yao and Sun, Lichao},
  journal={Advances in Neural Information Processing Systems},
  volume={37},
  pages={52127--52158},
  year={2024}
}

@inproceedings{liu2026attacking,
  title={Attacking Gray-Box Large Vision-Language Models with Adaptive SVD-Structured Adversarial Alignment},
  author={Liu, Daizong and Cai, Xiaowen and Dong, Junhao and Guo, Zhongliang and Qu, Xiaoye and Guan, Runwei and Fang, Xiang and Ye, Dengpan},
  booktitle={International Conference on Machine Learning},
  year={2026}
}

@inproceedings{liu2024unsupervised,
  title={Unsupervised domain adaptative temporal sentence localization with mutual information maximization},
  author={Liu, Daizong and Fang, Xiang and Qu, Xiaoye and Dong, Jianfeng and Yan, He and Yang, Yang and Zhou, Pan and Cheng, Yu},
  booktitle={Proceedings of the AAAI Conference on Artificial Intelligence},
  volume={38},
  number={4},
  pages={3567--3575},
  year={2024}
}

@inproceedings{liu2023hypotheses,
  title={Hypotheses tree building for one-shot temporal sentence localization},
  author={Liu, Daizong and Fang, Xiang and Zhou, Pan and Di, Xing and Lu, Weining and Cheng, Yu},
  booktitle={Proceedings of the AAAI Conference on Artificial Intelligence},
  volume={37},
  number={2},
  pages={1640--1648},
  year={2023}
}

@inproceedings{tang2024reparameterization,
  title={Reparameterization head for efficient multi-input networks},
  author={Tang, Keke and Zhao, Wenyu and Peng, Weilong and Fang, Xiang and Cui, Xiaodong and Zhu, Peican and Tian, Zhihong},
  booktitle={ICASSP 2024-2024 IEEE International Conference on Acoustics, Speech and Signal Processing (ICASSP)},
  pages={6190--6194},
  year={2024},
  organization={IEEE}
}

@article{xiong2024rethinking,
  title={Rethinking video sentence grounding from a tracking perspective with memory network and masked attention},
  author={Xiong, Zeyu and Liu, Daizong and Fang, Xiang and Qu, Xiaoye and Dong, Jianfeng and Zhu, Jiahao and Tang, Keke and Zhou, Pan},
  journal={IEEE Transactions on Multimedia},
  volume={26},
  pages={11204--11218},
  year={2024},
  publisher={IEEE}
}

@inproceedings{tang2025simplification,
  title={Simplification is all you need against out-of-distribution overconfidence},
  author={Tang, Keke and Hou, Chao and Peng, Weilong and Fang, Xiang and Wu, Zhize and Nie, Yongwei and Wang, Wenping and Tian, Zhihong},
  booktitle={Proceedings of the Computer Vision and Pattern Recognition Conference},
  pages={5030--5040},
  year={2025}
}

@article{cai2026towards,
  title={Towards building model/prompt-transferable attackers against large vision-language models},
  author={Cai, Xiaowen and Liu, Daizong and Qu, Xiaoye and Fang, Xiang and Dong, Jianfeng and Tang, Keke and Zhou, Pan and Sun, Lichao and Hu, Wei},
  journal={Advances in Neural Information Processing Systems},
  volume={38},
  pages={174022--174058},
  year={2026}
}

@article{yan2026fit,
  title={Fit the distribution: Cross-image/prompt adversarial attacks on multimodal large language models},
  author={Yan, Hai and Ma, Haijian and Cai, Xiaowen and Liu, Daizong and Yuan, Zenghui and Qu, Xiaoye and Dong, Jianfeng and Guan, Runwei and Fang, Xiang and He, Hongyang and others},
  journal={Advances in Neural Information Processing Systems},
  volume={38},
  pages={75204--75247},
  year={2026}
}

@inproceedings{liu2024towards,
  title={Towards robust temporal activity localization learning with noisy labels},
  author={Liu, Daizong and Qu, Xiaoye and Fang, Xiang and Dong, Jianfeng and Zhou, Pan and Nan, Guoshun and Tang, Keke and Fang, Wanlong and Cheng, Yu},
  booktitle={Proceedings of the 2024 Joint International Conference on Computational Linguistics, Language Resources and Evaluation (LREC-COLING 2024)},
  pages={16630--16642},
  year={2024}
}

@inproceedings{cai2025imperceptible,
  title={Imperceptible Beam-Sensitive Adversarial Attacks for LiDAR-based Object Detection in Autonomous Driving},
  author={Cai, Fuyao and Liu, Daizong and Fang, Xiang and Yu, Jixiang and Tang, Keke and Zhou, Pan},
  booktitle={2025 IEEE International Conference on Multimedia and Expo (ICME)},
  pages={1--6},
  year={2025},
  organization={IEEE}
}

@article{kuai2026dynamic,
  title={Dynamic Graph-enhanced Event Refinement for Temporal Sentence Grounding of Micro-moments},
  author={Kuai, Mingjin and Qin, You and Fang, Xiang and Ji, Wei and Zimmermann, Roger},
  journal={IEEE Transactions on Multimedia},
  year={2026},
  publisher={IEEE}
}

@inproceedings{fang2026towardsicml,
  title={Towards Understanding Modality Interaction in Multimodal Language Models via Partial Information Decomposition},
  author={Fang, Wanlong and Zhang, Tianle and Tao, Wen and Chan, Alvin},
  booktitle={International Conference on Machine Learning},
  year={2026}
}
\end{document}